\documentclass[lettersize,journal]{IEEEtran}
\usepackage{amsmath,amsfonts}
\usepackage{algorithmic}
\usepackage{array}
\usepackage[caption=false,font=normalsize,labelfont=sf,textfont=sf]{subfig}
\usepackage{textcomp}
\usepackage{stfloats}
\usepackage{url}
\usepackage{verbatim}
\usepackage{graphicx}
\def\BibTeX{{\rm B\kern-.05em{\sc i\kern-.025em b}\kern-.08em
    T\kern-.1667em\lower.7ex\hbox{E}\kern-.125emX}}
\usepackage{balance}
\usepackage{svg}
\usepackage{multirow}
\usepackage{booktabs}
\usepackage{cite}

\begin{document}

	\title{CD-CTFM: A Lightweight CNN-Transformer Network for Remote Sensing Cloud Detection Fusing Multiscale Features}

	\author{Wenxuan Ge, Xubing Yang, Li Zhang 
			\thanks{Wenxuan Ge, Xubing Yang, Li Zhang are with the College of
				Information Science and Technology, Nanjing 210037, China (gwx@njfu.edu.cn, xbyang@njfu.edu.cn, lizhang@njfu.edu.cn)
			}
			\thanks{This work is supported by Innovation and Entrepreneurship Training Program for college students in Jiangsu Province(No. 202210298120Y). 
			(Corresponding author: Li Zhang.)}
	}

	\markboth{Journal of \LaTeX\ Class Files,~Vol.~18, No.~9, September~2020}%
	{Wenxuan Ge, Li Zhang, \MakeLowercase{\textit{(et al.)}: Paper Title}}

	\maketitle

	\begin{abstract}
		Clouds in remote sensing images inevitably affect information extraction, which hinder the following analysis of satellite images. Hence, cloud detection is a necessary preprocessing procedure. However, the existing methods have numerous calculations and parameters. In this letter, a lightweight CNN-Transformer network, CD-CTFM, is proposed to solve the problem. CD-CTFM is based on encoder-decoder architecture and incorporates the attention mechanism. In the decoder part, we utilize a lightweight network combing CNN and Transformer as backbone, which is conducive to extract local and global features simultaneously. Moreover, a lightweight feature pyramid module is designed to fuse multiscale features with contextual information. In the decoder part, we integrate a lightweight channel-spatial attention module into each skip connection between encoder and decoder, extracting low-level features while suppressing irrelevant information without introducing many parameters. Finally, the proposed model is evaluated on two cloud datasets, 38-Cloud and MODIS. The results demonstrate that CD-CTFM achieves comparable accuracy as the state-of-art methods. At the same time, CD-CTFM outperforms state-of-art methods in terms of efficiency. 
	\end{abstract}


	\section{Introduction}
		\IEEEPARstart{R}{emote} sensing images collected from satellites are widely used in various fields, including land cover mapping, weather forecasting, and other fields \cite{RS-application}. However, optical remote sensing images are inevitably affected by cloud appearing on them, resulting in attenuation or loss of image information. Therefore, to improve the utilization of images with cloud, it is necessary to detect cloud before analyzing images. 
		
		Over the years, researchers have proposed a multitude of approaches from different perspectives. Traditional cloud detection methods can be broadly divided into two categories, threshold-based and machine-learning-based methods. Threshold-based methods are implemented using a threshold of different image parameters \cite{threshold}. Among threshold-based algorithms, the Function of mask (Fmask) is a widely used cloud detection approach \cite{Fmask}. 
		But threshold-based methods have poor universality in that thresholds vary as per the location. Machine-learning-based methods, such as decision tree, support vector machine, and Bayesian classification, identify cloud by learning from training data, which improves the performance of cloud detection \cite{threshold}. However, machine-learning-based methods are highly affected by the hand-craft features, which is highly dependent on expert knowledge and experience.
		
		With the advance in deep learning, methods based on convolutional neural networks have achieved great success in the field of computer vision, overwhelmingly surpassing traditional algorithms \cite{DL}. Scholars have applied deep learning to optical satellite imagery, including cloud detection. Francis \textit{et al.} proposed CloudFCN, which is based on fully convolutional networks, to extract multiscale spectral features of remote sensing images \cite{CloudFCN}. 
		Moreover, the attention mechanism was incorporated to improve accuracy \cite{CloudAttU, zhang}. 
		However, these cloud detection methods have a large number of parameters and computations. To alleviate this problem, Yao \textit{et al.} designed a light-weight channel attention module in their CD-AttDLV3+, which was based on DeepLabV3+ architecture, to strengthen the learning of significant channels and reduce computations \cite{CD-AttDLV3+}. 
		However, there are still two problems in these cloud detection methods. On the one hand, existing methods only utilize the stack of convolutional layers to obtain local spatial features, ignoring global semantic information of image. On the other hand, existing models with attention mechanism still have much room for improvement in both performance and efficiency.
		
		In this letter, we propose a lightweight encoder-decoder architecture with attention mechanism, CD-CTFM, for cloud detection. To summarize, the main contributions of this work are listed as follows: 1) CD-CTFM utilizes a lightweight CNN-Transformer network as backbone. Compared to traditional Convolutional Neural Network (CNN), CNN-Transformer network can extract both local and global features, which is conducive to improve accuracy and reduce computations. 2) A lightweight feature pyramid module enables CD-CTFM to fuse multiscale features with global features. 3) CD-CTFM designs a lightweight channel-spatial attention module to suppress invalid information and highlight discriminative features in a lightweight way. Finally, two cloud datasets, 38-Cloud and MODIS, were used to to validate the effectiveness of our proposed model.

	\section{Methodology}
		\subsection{Overview of CD-CTFM}
			\begin{figure*}[!t]
				\centering
				\includegraphics[width=0.85\textwidth]{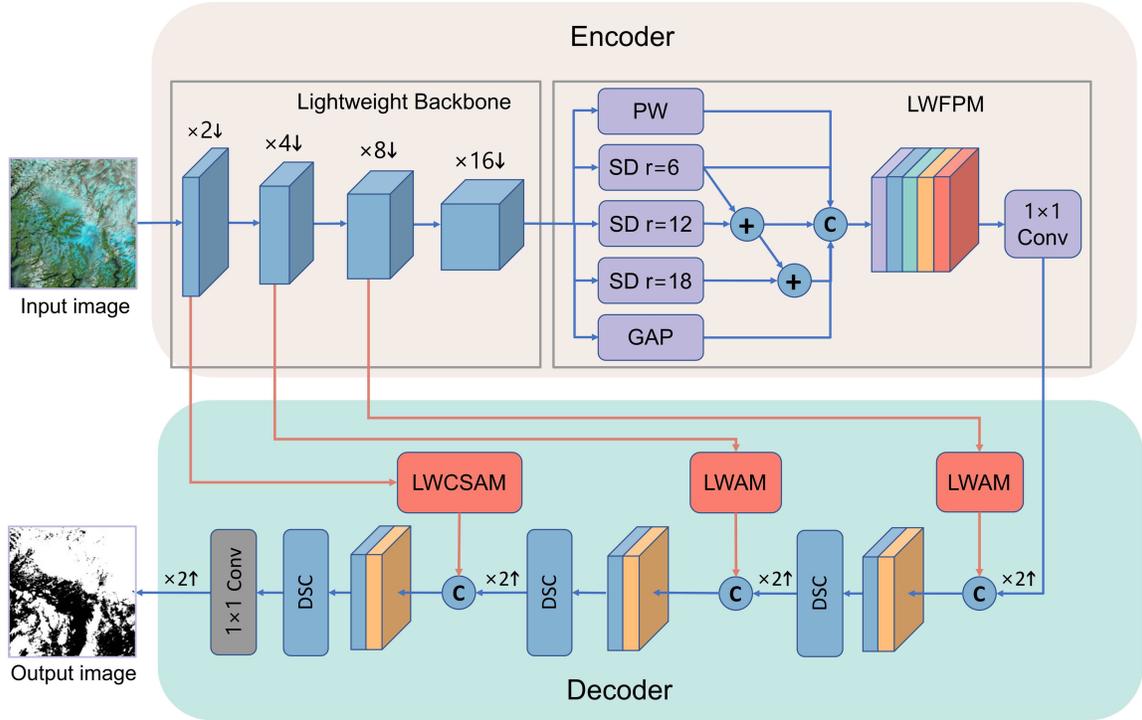}	
				\caption{Overall framework of CD-CTFM. The encoder contains a lightweight backbone and LWFPM while the decoder is based on depthwise separable convolutions (DSC). The LWAM filters information propagated through the skip connections.}
				\label{CD-CTFM}
			\end{figure*}
		
			\begin{figure}[!t]
				\centering
				\includegraphics[width=0.45\textwidth]{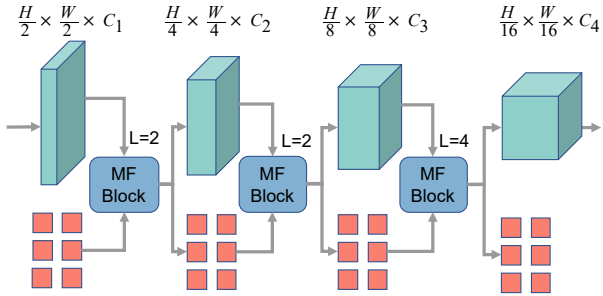}
				\caption{The detail of the lightweight backbone.}
				\label{MF}
			\end{figure}
		
			The CD-CTFM exploits the encoder-decoder architecture, as shown in Fig \ref{CD-CTFM}. In the encoder part, we deploy a hybrid network of CNN and Transformer as backbone, as shown in Fig \ref{MF}, which is based on Mobile-Former block \cite{MF}. A convolutional layer with 3$\times$3 kernel size is first adopted in the backbone to adjust the number of channels and downsample the feature maps. Then the feature maps are forwarded to a sequence of Mobile-Former block, which contains four parts. The Mobile sub-block consists of a bottleneck block, and the former sub-block is the stacking of multi-head attention and feed-forward network. Besides, Mobile2Former and Former2Mobile are lightweight cross attention gates utilized to forward local and global features. In the backbone, local spatial features and global context information are extracted simultaneously and fused bidirectionally, resulting in more representation power. Followed by the backbone is the lightweight feature pyramid module (LWFPM), fusing high-level multiscale features with global interaction information. In the decoder part, deep features coming from LWFPM are up-sampled layer by layer. To compensate for the attenuated or lost low-level features, we introduce skip connection and integrated a lightweight channel-spatial attention module (LWAM) into each skip connection, which suppress irrelevant low-level information and highlight discriminative features.

		\subsection{Lightweight Feature Pyramid Module (LWFPM)}
			\begin{figure}[!t]
				\centering
				\includegraphics[width=0.45\textwidth]{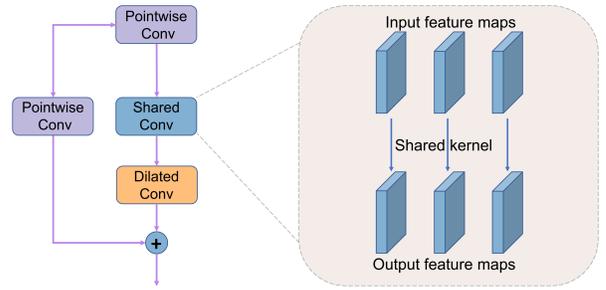}
				\caption{The structure of SD block.}
				\label{SD}
			\end{figure}
			
			To further extract features propagated through backbone, we designed LWFPM, which consists of five parallel paths (see Fig \ref{CD-CTFM}). Except for one global average pooling (GAP) path and one pointwise convolution (PW) path, the rest of the paths all consist of a novel shared and dilation block (SDblock), of which the dilated rate is 6, 12, and 18, respectively. As shown in Fig \ref{SD}, SDblock, based on residual structure, contains a PW, a shared convolution (SC) and a dilated convolution (DC). The effect of the PW is dimensionality reduction while the purpose of SC is deep features extraction, both of which are conducive to decrease parameters and computations. To fuse high-level multiscale features with context information, dilated convolutions with different dilated rate are added. Inspired by ESPNet \cite{ESP}, LWFPM deploys hierarchical feature fusion to solve the gridding effect caused by dilated convolutions without introducing extra parameters.

		\subsection{Lightweight Channel-Spatial Attention Module (LWAM)}
			\begin{figure}[!t]
				\centering
				\includegraphics[width=0.45\textwidth]{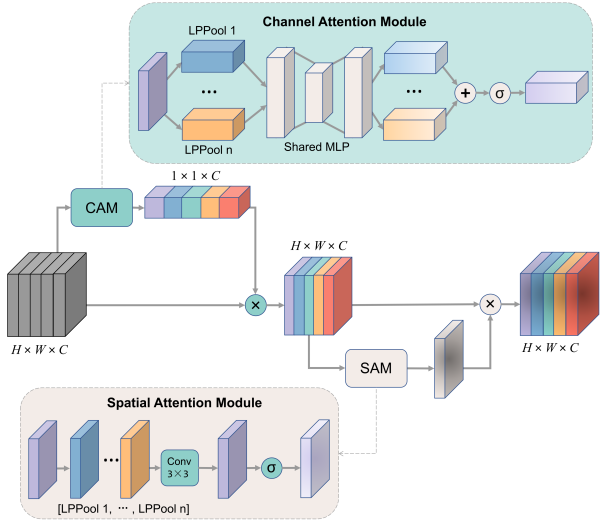}
				\caption{The detail of Lightweight Channel-Spatial Attention Module}
				\label{LWAM}
			\end{figure}
			By the skip connections, CD-CTFM has more information to detect cloud. However, some feature channels contains too much invalid or irrelevant information, affecting the accuracy of the model. Similarly, in each feature map, each pixel contributes differently to the semantic segmentation task. To alleviate this problem, a lightweight channel-spatial attention module is proposed, which is based on channel attention module (CAM) and spatial attention module (SAM) \cite{CBAM}. As shown in Fig \ref{LWAM}, the CAM and SAM are cascaded directly. In CAM, the input feature maps are processed by using LP pooling \cite{LP}. In this letter, the number of pooling layers and the p-value of each path is set as 2, 1, and 2. After the pooling operations, we forward both feature maps to a shared multi-layer perceptrons (MLP), generating the attention map. Then the feature maps are summed at the element level. Similarly, the input feature maps are pooled in the same way as CAM. Then the outputs of pooling operations are concatenated, followed by a convolution layer with 3$\times$3 kernel size.

	\section{Experiments and Results}
		\subsection{Experimental Settings}
			\subsubsection{Datasets}
				To train and test the CD-CTFM, we conduct experiments on two cloud detection datasets: 38-Cloud and MODIS datasets.
				
				The 38-Cloud dataset is first released in \cite{38-cloud-2}. The dataset contains 18 LandSat-8 images for training and 20 images for testing. Due to the large size of images, it is difficult to directly use theses images as inputs. Therefore, each image is cropped into 384 × 384 non-overlapping patches. After cropping, the training set contains 8400 patches while the test set contains 9201 patches. Each patch has 4 corresponding spectral channels: Red (band 4), Green (band 3), Blue (band 2), and Near Infrared (band 5).
				
				MODIS dataset contains 1422 remote sensing images, which are separated into 1272 training images and 150 test images \cite{MODIS}. After cropping them into 512 x 512 patches, the training set and test set contain 19080 and 2250 non-overlapping patches, respectively. Each patch consists of ten spectral channels: band 1, 3, 4, 18, 20, 23, 28, 29, 31, and 32.
				
			\subsubsection{Evaluation Metrics}
				In the experiment, the ground truths and prediction results are divided into cloud and non-cloud classes at pixel level. The performance of CD-CTFM is evaluated by five quantitative metrics, including mean intersection-over-union (mIoU), precision, recall, F1-score, and overall accuracy (OA). These metrics are defined as follows:
				\begin{align}
					\text{mIoU} &= \frac{\text{TP}}{\text{TP + FN + FP}}  \\
					\text{Precision} &= \frac{\text{TP}}{\text{TP + FP}} \\
					\text{Recall} &= \frac{\text{TP}}{\text{TP + FN}} \\
					\text{F1-score} &= \frac{\text{2}\times\text{TP}}{\text{2}\times\text{TP + FP + FN}}  \\
					\text{OA} &= \frac{\text{TP + TN}}{\text{TP + TN + FP + FN}}
				\end{align}
				where TP, TN, FP, and FN represent the total number of true positive, true negative, false positive and false negative pixels, respectively. Besides, we utilize two metrics, including giga floating-point operations per second (GFLOPS) and the number of model parameters, to measure the efficiency of models.

			\subsubsection{Implementation Details}
				Our experiments are based on the Pytorch framework and Ubuntu 20.04 equipped with an NVIDIA 3090 24G GPU. All models are optimized by the stochastic gradient descent (SGD) algorithm and the learning rate decays from 0.001 to 0. Besides, the batch sizes, momentum, and epochs are 32, 0.9, and 50, respectively.
			
		\subsection{Ablation Studies}
			\begin{table*}
				\begin{center}
					\caption{Ablation Study of LWAM and LWFPM on 38-Cloud Test Set}
					\label{ablation result}
					\begin{tabular}{ c | c c c c c | c c }
						\toprule[1pt]
						\multirow{2}{*}{Method} & \multicolumn{5}{c|}{Performance}& \multicolumn{2}{c}{Efficiency}\\
						\cline{2-8}
						\rule{0pt}{8pt}
						& mIoU & Precision & Recall & $F_1$-Score & OA & params (M) & GFLOPS\\
						\midrule[0.4pt]
						Backbone  			& 83.07	& 88.66	& 90.26	& 89.45	& 95.03	& 2.31 & 0.31 \\ 
						Backbone+LWFPM 		& 83.72 & 90.07	& 89.63	& 89.85	& 95.19	& 2.77 & 0.39 \\
						Backbone+LWAM 	  	& 83.17	& 90.32	& 88.76	& 89.53	& 94.99	& 2.31 & 0.32 \\
						Backbone+LWFPM+LWAM & 84.13	& 91.09	& 89.22	& 90.15	& 95.45	& 2.77 & 0.40 \\
						\bottomrule[1pt]
					\end{tabular}
				\end{center}
			\end{table*}
		
			In this section, we utilize 38-Cloud dataset to verify the effectiveness of the key modules in CD-CTFM, which are LWFPM and LWAM. To begin with, we train and test the CD-CTFM without LWFPM and LWAM in order to evaluate the performance and efficiency of backbone. Subsequently, we conducted experiments on backbone with LWFPM and LWAM, respectively. In particular, in backbone+LWFPM method, the low-level feature maps from encoder are sent directly to decoder. Also, in backbone+LWAM method, the LWFPM module is replaced with a pointwise convolution. Table~\ref{ablation result} shows the performance and efficiency of CD-CTFM under different settings. Obviously, when equipped with LWFPM and LWAM, the model obtain the best evaluation on performance. The results of this section manifest that both LWFPM and LWAM improve the performance a lot while decrease the efficiency a little bit. In particular, LWFPM and LWAM help a lot to increase mIoU.
			
		\subsection{Comparison with Other Methods}
			\begin{table*}
				\begin{center}
					\caption{Cloud Detection Performance Comparisons and Efficiency Comparisons on 38-Cloud and MODIS Datasets}
					\label{comparison result}
					\begin{tabular}{ c c | c c c c c | c c }
						\toprule[1pt]
						\multirow{2}{*}{Dataset} & \multirow{2}{*}{Method} & \multicolumn{5}{c|}{Performance}& \multicolumn{2}{c}{Efficiency}\\
						\cline{3-9}
						\rule{0pt}{8pt}
						&  & mIoU & Precision & Recall & $F_1$-Score & OA & params (M) & GFLOPS\\	
						\midrule[0.4pt]
						\multirow{7}{*}{38-Cloud}	& DeepLabv3+  & 81.64 & 87.72 & 88.36 &	88.04 &	95.03 & 41.04 & 13.24 \\
													& UNet 		  & 83.82 & 89.68 & 89.85 &	89.76 & 95.75 & 31.05 & 42.02 \\
													& CloudFCN 	  & 83.31 & 88.81 & 89.61 &	89.21 &	95.66 & 11.44 & 40.53 \\
													& CloudAttU   & 84.73 & 90.62 & 89.95 & 90.28 & 95.92 &	40.45 & 64.21 \\
													& MobileNetV2 & 82.37 &	90.97 & 87.31 &	89.10 &	94.90 & 3.31  & 5.65  \\
													& CD-AttDLV3+ & 81.24 & 88.85 &	87.58 &	88.21 & 94.49 &	3.54  & 1.33  \\
													& CD-CTFM     & 84.13 & 91.09 & 89.22 & 90.15 & 95.45 & 2.77  & 0.40  \\
						\hline
						\rule{0pt}{8pt}
						\multirow{7}{*}{MODIS}	& DeepLabv3+  & 81.57 & 90.13 &	88.62 &	88.67 &	92.53 &	42.15 &	13.26 \\
												& UNet 		  & 86.01 &	92.82 &	91.44 &	91.65 &	94.67 &	31.05 &	42.19 \\
												& CloudFCN 	  &	85.65 &	92.04 &	91.79 &	91.45 &	94.36 &	11.47 &	42.04 \\
												& CloudAttU   & 86.04 &	92.93 &	91.42 &	91.66 &	94.71 &	40.45 &	64.39 \\
												& MobileNetV2 & 82.19 &	90.59 &	88.75 & 89.14 &	92.87 &	3.32  &	5.67  \\
												& CD-AttDLV3+ & 82.28 &	90.87 &	88.80 &	89.14 &	92.93 &	3.39  &	1.36  \\
												& CD-CTFM     & 84.77 &	91.96 &	90.52 &	90.90 &	94.17 &	2.77  &	0.42  \\
						\bottomrule[1pt]
					
					\end{tabular}
				\end{center}
			\end{table*}
			\begin{figure*}[!t]
				\centering
				\includegraphics[width=0.9\textwidth]{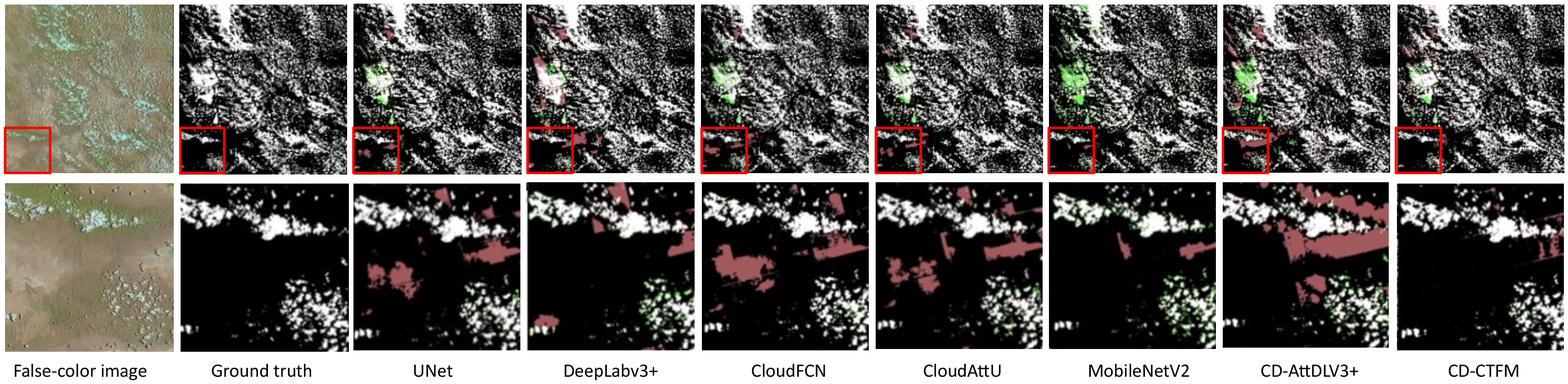}
				\caption{Comparison between the results of different methods in 38-Cloud dataset. White area represents cloud, black area represents non-cloud, red area represents false-positive detection and green area represents false-negative detection.}
				\label{result-L}
			\end{figure*}
			\begin{figure*}[!t]
				\centering
				\includegraphics[width=0.9\textwidth]{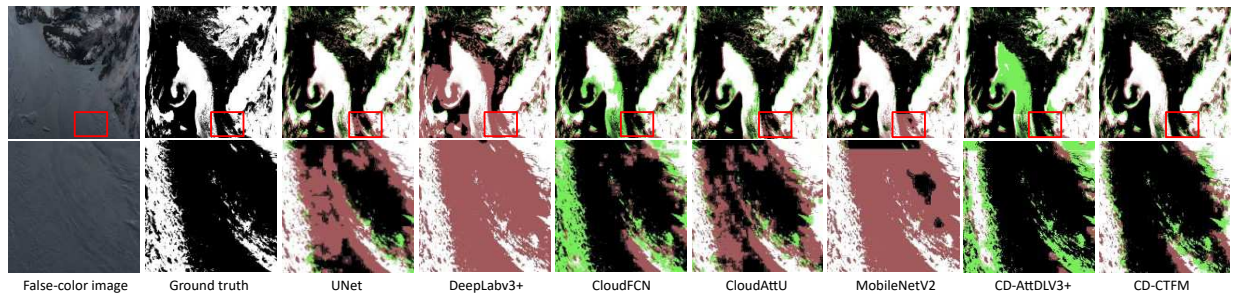}
				\caption{Comparison between the results of different methods in MODIS dataset. White area represents cloud, black area represents non-cloud, red area represents false-positive detection and green area represents false-negative detection. }
				\label{result-M}
			\end{figure*}
			To further validate the effectiveness of our proposed model, we compare CD-CTFM with two common semantic segmentation model (UNet \cite{u} and DeepLabV3+ \cite{DLab3+}), two cloud detection model (CloudFCN and CloudAttU), one common lightweight model (MobileNetV2 \cite{mv2}), and one lightweight cloud detection model (CD-AttDLV3+) on MODIS and 38-Cloud datasets. The quantitative results are shown in Table~\ref{comparison result}, from which we can see that CD-CTFM outperforms, in terms of efficiency, MobileNetV2 and CD-AttDLV3+. In 38-Cloud dataset, the mIoU index of CD-CTFM is 0.6$\%$ lower than CloudAttU, which has the highest value in mIoU, but the parameters and computations of CloudAttU are very large, which are 14.6 and 160.5 times as large as CD-CTFM, respectively. Compared to other methods, CD-CTFM achieves better or similar performance. The results are similar in MODIS datasets. Fig \ref{result-L} and Fig \ref{result-M} demonstrates the qualitative results of compared methods on 38-Cloud and MODIS datasets. Fig \ref{result-L} contains flat lands with complicated texture and broken clouds. CD-CTFM can not only clearly detect correct cloud regions, but also exclude surface with complex texture, which are like clouds in terms of color and shape. Fig \ref{result-M} shows snow and ice surface with thin clouds. CD-CTFM is able to identify
			ice and snow surfaces with complex textures, benefiting from its powerful representation learning capability. In addition, CD-CTFM can distinguish cloud from the cloud-like ice and snow with the help of global background scenes.

	\section{Conclusion}
		In this letter, we propose a lightweight cloud detection model, CD-CTFM, to reduce parameters and computations. To enable our model to extract local and global feature simultaneously, we utilize a lightweight CNN-Transformer network as the backbone. Also, we design a lightweight feature pyramid module to fuse multiscale features with global context information. Then a lightweight channel-spatial attention module is integrated between the encoder and the decoder through skip connection. Experimental results on 38-Cloud and MODIS datasets demonstrate that CD-CTFM achieves better or similar performance while decreasing parameters and computations, compared to state-of-art methods. We will introduce knowledge distillation and model compression techniques to further improve the accuracy and reduce computations of our model. 

\balance

\bibliographystyle{IEEEtran}
\bibliography{paper}

\end{document}